\documentclass{article}

\PassOptionsToPackage{numbers, compress}{natbib}

\usepackage[preprint]{neurips_2024}

\usepackage{times}
\usepackage{booktabs}
\usepackage{latexsym}
\usepackage{graphicx}
\usepackage{multirow}
\usepackage{multicol}
\usepackage{hyperref}
\usepackage{listings}
\usepackage{amsmath}
\usepackage[most]{tcolorbox}
\usepackage{float}
\usepackage{xspace}
\usepackage{enumitem}
\usepackage[T1]{fontenc}
\usepackage{xcolor}
\usepackage{microtype}
\usepackage{inconsolata}
\usepackage[utf8]{inputenc}
\usepackage{caption}
\newcommand{\shortname}{{CIBench}}

\title{\shortname: Evaluating Your LLMs with a Code Interpreter Plugin}

\author{Chuyu Zhang$^{1,2,*}$, ,Songyang Zhang$^{1,}$\thanks{Equal contribution. Code is available at \href{https://github.com/open-compass/CIBench}{https://github.com/open-compass/CIBench}}, Yingfan Hu$^{*}$, Haowen Shen, Kuikun Liu$^{1}$,\\ \textbf{Zerun Ma}$^{1}$,
 \textbf{Fengzhe Zhou}$^{1}$, \textbf{Wenwei Zhang}$^{1}$, \textbf{Xuming He}$^{2}$, \textbf{Dahua Lin}$^{1}$, \textbf{Kai Chen}$^{1,\dagger}$ \\
{\normalsize $^1$Shanghai Artificial Intelligence Laboratory} \quad
{\normalsize $^2$ShanghaiTech University}  \\
{\tt\small \{zhangsongyang\}@pjlab.org.cn}
\vspace{-2em}
}

\begin{document}

\maketitle

\begin{abstract}

While LLM-Based agents, which use external tools to solve complex problems, have made significant progress, benchmarking their ability is challenging, thereby hindering a clear understanding of their limitations. In this paper, we propose an interactive evaluation framework, named~\shortname, to comprehensively assess LLMs' ability to utilize code interpreters for data science tasks. Our evaluation framework includes an evaluation dataset and two evaluation modes. The evaluation dataset is constructed using an LLM-human cooperative approach and simulates an authentic workflow by leveraging consecutive and interactive IPython sessions. The two evaluation modes assess LLMs' ability with and without human assistance. We conduct extensive experiments to analyze the ability of 24 LLMs on \shortname~and provide valuable insights for future LLMs in code interpreter utilization.

\end{abstract}

\section{Introduction}
\vspace{-0.3em}

Empowered by the emerging abilities of Large Language Models (LLMs), numerous agents have surfaced to harness these generalist models for utilizing external tools, such as web browsing, document retrieval, and code interpretation, in tackling complex real-world problems. Notable examples include ToolLLM~\cite{qin2023toolllm}, LangChain~\cite{Chase_LangChain_2022}, AutoGPT~\cite{Significant_Gravitas_AutoGPT}, and QwenAgent~\cite{bai2023qwen}. Specifically, agents equipped with a code interpreter leverage the advanced programming skills of LLMs through a natural language interface, thereby facilitating the creation of workflows that are both effortless and efficient. \textit{However, assessing the agents' capability to generate executable code and effectively use code interpreters remains a challenge, hindering a clear understanding of their limitations.}

In this study, we focus on assessing the proficiency of LLMs in leveraging code interpreters to address data science problems across several distinct domains, like data analysis, visualization, and machine learning. These tasks necessitate that LLMs exhibit advanced capabilities in instruction following, reasoning, and programming. 
Existing benchmarks, including GSM8K~\cite{cobbe2021training}, MathBench~\cite{liu2024mathbench} and HumanEval~\cite{chen2021evaluating}, primarily assess the models' abilities to solve mathematical or coding problems solely. These benchmarks, while valuable for measuring specific competencies, fail to fully represent the LLMs' aptitude for building complex workflows involving code interpreters in practical applications. 
More recent efforts, such as MINT~\cite{wang2023mint}, QwenAgent~\cite{bai2023qwen}, CodeGen~\cite{nijkamp2022codegen}, and DS-1000~\cite{lai2023ds}, aim to bridge the existing gap by constructing novel benchmarks, particularly tailored to mathematical problems and data analysis tasks.
Despite considerable advancements, they either focus on single-turn question assessments or have a limited scope in data science. 
The substantial insights they provided inadequately reflect the LLM's ability to invoke a code interpreter to solve real-world data science problems, which typically require consecutive sessions.

\begin{figure}[!t]
    \centering
    \includegraphics[scale=0.2]{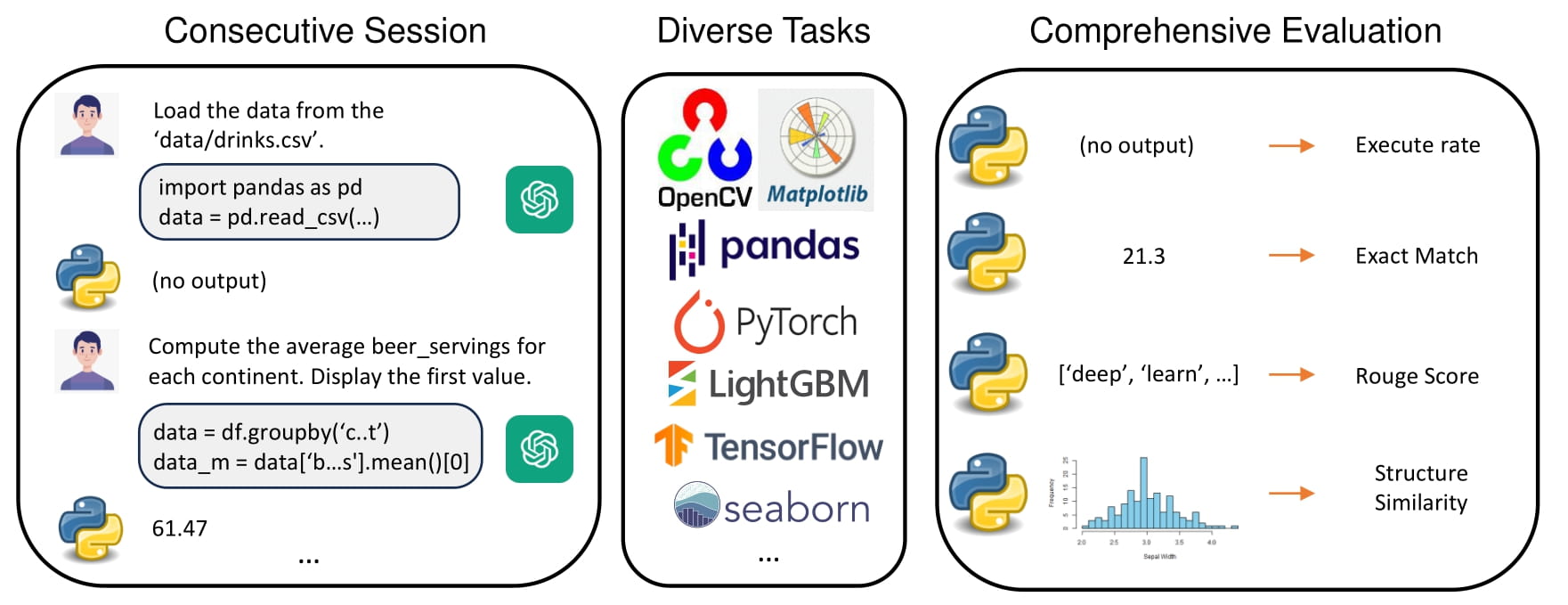}
    \caption{Features of our benchmark. Our benchmark consists of interactive sessions, diverse tasks covering various Python modules, and comprehensive evaluations (The tool-call rate is not displayed).}
    \label{fig:proper}
    \vspace{-1.5em}
\end{figure}

To address these shortcomings, as depicted in Fig.\ref{fig:proper}, we introduce a novel evaluation framework that encompasses a benchmark with consecutive and diverse tasks, along with comprehensive assessment protocols. This framework aims to provide a thorough evaluation of LLMs' ability to use code interpreters. Notably, the benchmark employs a distinctive \textit{LLM-human cooperative approach}, as depicted in Fig.~\ref{fig:main}. It simulates authentic workflow scenarios using \textit{interactive IPython} sessions with sequential, interconnected questions focused on popular Python modules such as \textit{Matplotlib}, \textit{Pandas}, and \textit{PyTorch}. Additionally, we utilize two evaluation modes and a series of fine-grained metrics for various outputs. The statistics of our benchmark are shown in Tab.\ref{existing_work}. 




\begin{table}[!t]
\Huge
\centering
\caption{Comparison of different datasets. "Multi-turn" denotes that LLMs can attempt to solve tasks through multiple trials. "Code interpreter" means whether LLM needs to call the code interpreter to solve the problem. "Consecutive" indicates that each task consists of a series of sequential questions rather than a single question.}
\label{existing_work}
\resizebox{\textwidth}{!}{
\begin{tabular}{c|ccccc}
\toprule
Dataset & Topic                              & Size                      &Multi-turn  & Code Interpreter & Consecutive   \\ \midrule
DS-1000 & Data Science                       & 1000 questions            & No & No & No          \\
MINT    & Math, Reasoning                    & 586 questions             & Yes & Yes & No         \\
CodeGen & Math, Data Science, etc & 115 tasks                 & No & No & Yes         \\
QwenAgent & Math, Data visualization & 295 questions                 & No & Yes  & No        \\ \midrule
\shortname & Data Science                       & 234 tasks, 1900+ questions & Yes & Yes & Yes        \\ \bottomrule
\end{tabular}}
\vspace{-0.5em}
\end{table}

Specifically, we build the evaluation dataset by initially identifying ten highly prevalent Python libraries within the domain of data science. Then, we prompt advanced LLM, such as GPT-4, to generate instructions and code snippets within Jupyter Notebook format. Each notebook is structured to contain a sequence of 10 to 15 progressive steps, with increasing levels of complexity.
To mitigate any inherent biases or limitations in the LLM-generated content, we employ human experts to generate template tasks based on the common patterns observed in the LLM-generated tasks and online resources. Those tasks can incorporate multiple interchangeable datasets for evaluation. Those designs ensure that the benchmark encapsulates both diversity and quality, thereby offering a comprehensive and balanced assessment of code interpreter capabilities.  


To thoroughly assess the LLMs' performance on our benchmark, we have instituted two distinct evaluation modes: the \textit{end-to-end mode} and the \textit{oracle mode}.
In the end-to-end mode, LLMs are tasked with a holistic problem-solving process where they must reason through given instructions and generate corresponding code. This requires them to iteratively refine their output based on feedback from the code interpreter, as they attempt to solve multiple consecutive questions that build upon one another.
Additionally, the oracle mode simulates guided learning by providing the LLM with the correct code snippet when it fails. This immediate feedback mimics human guidance and equips the model to use this accurate example for tackling subsequent tasks in the same context. 
Furthermore, we introduce two types of metrics: the process-oriented (\textit{i.e., tool-call rate, executable rate}) and output-oriented (\textit{i.e., numeric accuracy, text score, visualization score}), to provide a comprehensive analysis of the model’s performance.




Based on our evaluation framework, we conduct extensive experiments and analysis using 19 LLMs. The results indicate that open-sourced LLMs struggle to utilize PyTorch- and TensorFlow-like modules (Fig.\ref{fig:radar}), and the best-open-sourced LLMs lag behind GPT-4 by 10.0\% (Tab.\ref{tab:main_results}). In summary, our contributions are three-fold:

\noindent$\bullet$ We build a new benchmark for agents with code interpreters using an LLM-human cooperative method. It consists of interactive IPython sessions with interconnected questions on key data science libraries, simulating interactive problem-solving scenarios in practical workflows.

\noindent$\bullet$ We devise unique assessment strategies involving both end-to-end and oracle modes. We also introduce several evaluation metrics to assess various outputs, offering a comprehensive gauge of LLMs' coding prowess within the benchmark.
    
\noindent$\bullet$ We conduct thorough experiments with 24 LLMs to analyze their performance on our benchmark. The results indicate that open-sourced LLMs are inferior to GPT-4 by a large margin and perform poorly in the modeling category modules.

\section{Related Works}
\vspace{-0.5em}
\shortname~is an evaluation framework that assesses LLMs'~\cite{touvron2023llama2,bai2023qwen,deepseekai2024deepseek,chiang2023vicuna,cai2024internlm2} ability to utilize external code interpreters for solving data science tasks. Therefore, we focus on presenting work related to invoking code interpreters and benchmarks related to data science.

\vspace{-0.5em}
\subsection{Model with Plugins}
\vspace{-0.5em}
LLM-based agents use external tools via APIs to solve complex tasks and have been regarded as a promising direction~\cite{Chase_LangChain_2022,qin2023toolllm,Significant_Gravitas_AutoGPT,schick2023toolformer,hong2023metagpt,wu2023autogen}.
Specifically, \cite{li2023camel,hong2023metagpt,qian2023communicative} develop efficient workflows to coordinate multi-agent systems for automatic programming. \cite{schick2023toolformer,Chase_LangChain_2022,qin2023toolllm,gao2023pal} equip LLMs with external tools, such as search engines, calculators, and code interpreters, to augment LLMs' problem-solving ability. Among these tools, the code interpreter can promote LLMs' reasoning and coding ability and has gradually gained attention in works like~\cite{ying2024internlm,zhou2023solving,zhuang2023toolqa}.

In contrast, we aim to develop an evaluation framework to benchmark LLMs' ability with code interpreters for solving data science tasks, providing insights for future work to promote LLMs for better realistic utilization.
\vspace{-0.5em}
\subsection{Related Benchmarks}
\vspace{-0.5em}
Various benchmarks~\cite{cobbe2021training,chen2021evaluating,qin2023toolllm,bai2023qwen} have been proposed to measure LLMs' reasoning, coding, and tool utilization ability. Classic benchmarks, such as GSM8K~\cite{cobbe2021training}, MATH~\cite{hendrycks2021measuring}, HumanEval~\cite{chen2021evaluating}, and MBPP~\cite{austin2021program}, focus on evaluating the mathematical reasoning or code capability of LLMs. ToolBench~\cite{qin2023toolllm} and MS-Agent Bench~\cite{li2023modelscopeagent} aim to evaluate LLMs' capability in effectively utilizing various tools and generating accurate and contextually appropriate responses. ~\cite{wang2023mint} introduce MINT, a benchmark that evaluates LLMs’ ability to solve challenging tasks with multi-turn interactions by using tools and leveraging natural language feedback. However, the above benchmarks cannot measure LLMs' ability in data science tasks, which require instruction following, coding, and tool utilization abilities. To address this gap, ~\cite{nijkamp2022codegen} introduces a multi-turn programming benchmark consisting of 115 diverse problems, covering topics such as mathematics, algorithms, and data science. QwenAgent~\cite{bai2023qwen} introduces a benchmark for data science, focusing mainly on mathematical problems and data visualization. ~\cite{lai2023ds} introduce DS-1000, a code generation benchmark with a thousand data science problems.  Meanwhile, \cite{hu2024infiagent} introduces DABench, which evaluates various concepts with individual questions assigned to each dataset. However, this approach does not effectively mimic practical scenarios where interconnected questions are raised.

In contrast to the above works, \shortname~ simulates a real-world data science workflow by leveraging sequential interactive IPython sessions and covers most concepts in data science by including commonly used Python modules. Furthermore, we devise two evaluation modes and five metrics to holistically evaluate LLMs' abilities.
\definecolor{lightgray}{gray}{0.95}
\definecolor{codegreen}{rgb}{0,0.6,0}
\definecolor{codegray}{rgb}{0.5,0.5,0.5}
\definecolor{codepurple}{rgb}{0.58,0,0.82}
\lstdefinestyle{prompt}{
    basicstyle=\ttfamily\fontsize{7pt}{8pt}\selectfont,
    frame=none,
    breaklines=true,
    backgroundcolor=\color{lightgray},
    breakatwhitespace=true,
    breakindent=0pt,
    escapeinside={(*@}{@*)},
    numbers=none,
    numbersep=5pt,
    xleftmargin=5pt,
}
\lstdefinestyle{code}{
    language=Python,
    basicstyle=\ttfamily\fontsize{7pt}{8pt}\selectfont,
    keywordstyle=\color{magenta},
    commentstyle=\color{codegreen},
    stringstyle=\color{codepurple},
    showstringspaces=false，
    breaklines=true,
    backgroundcolor=\color{lightgray},
    breakatwhitespace=true,
    breakindent=0pt,
    numbers=none,
    numbersep=5pt,
    xleftmargin=5pt,
}
\tcbset{
  aibox/.style={
    top=10pt,
    colback=white,
    colframe=black,
    colbacktitle=black,
    enhanced,
    center,
    attach boxed title to top left={yshift=-0.1in,xshift=0.15in},
    boxed title style={boxrule=0pt,colframe=white,},
  }
}
\newtcolorbox{AIbox}[2][]{aibox, title=#2,#1}

\section{CIBench}\label{sec:cibench}
\vspace{-0.5em}

\begin{figure*}[!t]
    \centering
    \includegraphics[width=1.0\textwidth]{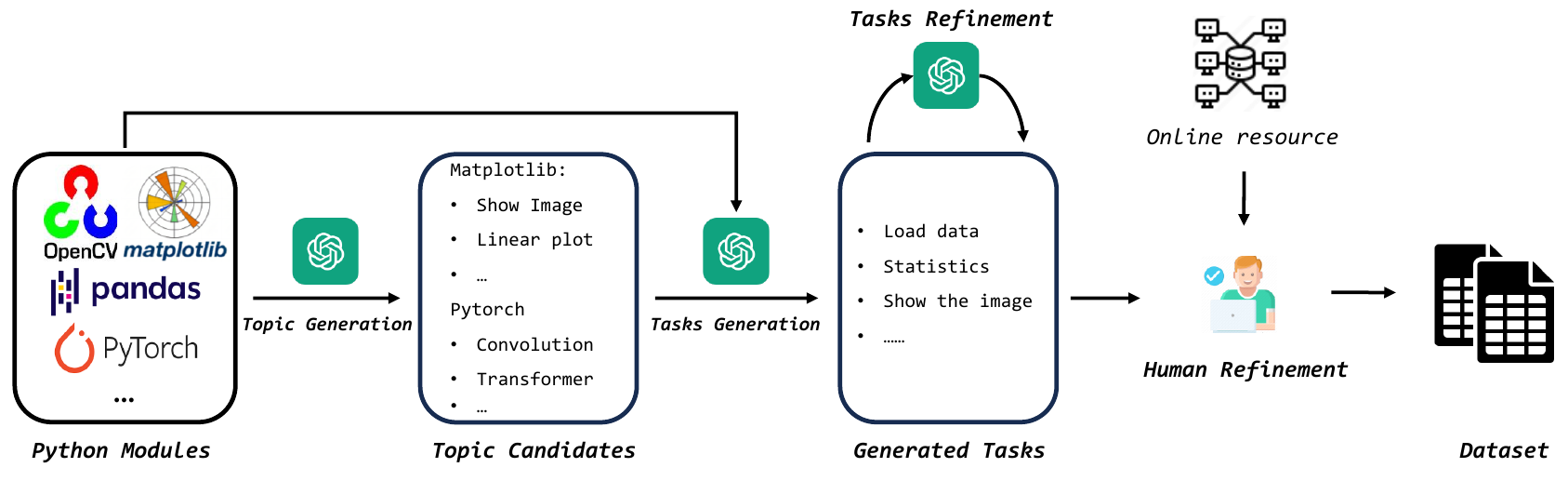}
    \vspace{-1em}
    \caption{Overview of \shortname. \shortname~first selects Python modules to generate candidate topics and then generates tasks based on these modules and the selected topic. Additionally, humans are engaged to generate new tasks to ensure diversity and filter out incorrect questions to enhance quality. }
    \label{fig:main}
    \vspace{-0.5em}
\end{figure*}

To benchmark LLM's ability to leverage code interpreters for addressing data science problems, we propose a novel evaluation framework (Fig.\ref{fig:main}), which comprises a diverse evaluation dataset and two newly devised evaluation modes. The evaluation dataset is generated through an LLM-human cooperative approach and simulates authentic workflow scenarios for solving sequential and interconnected tasks.
Given the evaluation dataset, we adhere to the ReAct protocol~\cite{yao2023react} to generate reasoning traces and invoke code interpreters alternately. And, we allow LLMs to attempt to solve tasks multiple times, enabling exploration of their self-debugging capabilities based on feedback from the code interpreter. Finally, we propose two evaluation modes: the end-to-end mode and the oracle mode, to comprehensively measure LLM's ability with and without human interaction.

In the following sections, we will detail the construction of the dataset in Sec.\ref{sec:Evaluation Dataset} and the evaluation modes and metrics in Sec.\ref{sec:Metrics}.

\begin{table*}[!t]
\centering{
\caption{Selected Python modules and their categories.}
\label{tab:python_modules}
\begin{tabular}{lr}
\toprule
\textbf{Category} & \textbf{Python Modules}\\
\midrule
Data Cleaning and Manipulation & Pandas \\
Data Visualization & Matplotlib, Seaborn \\
Modeling  & Scikit-learn, PyTorch, TensorFlow, LightGBM \\
Natural Language Processing & NLTK \\
Image Processing & OpenCV-Python\\
Mathematics and Statistics & SciPy\\
\bottomrule
\end{tabular}}
\vspace{-1em}
\end{table*}

\begin{figure*}[!t] 
\begin{AIbox}{Task Generation}
{\color{blue}\bf Prompt:} \\
{
Please create jupyter notebook experiment based on Python module \{\}. Please follow these rules:\\
1. The experiment should be conducted in a jupyter notebook manner, but use the markdown format. \\
2. The experiment should only use Python code.\\
3. The experiment has around 10-15 continuous steps, from the easiest to the hardest.\\
4. The step description should be concise.\\
5. The step description should be precise and contain exact parameter names and values to instruct.\\
6. Each step requires Python code to solve and the executed result should be the numeric answer, structured output, or visualized result.\\
7. Please use `matplotlib' to visualize if necessary.\\
8. DO NOT have any steps to save or write any output files.\\
9. Please provide an input data file with an external link.\\
The experiment topic is \{\}. You should generate the experiment file without any other statements.
}
\end{AIbox}
\vspace{-1em}
\caption{An example prompt of task generation.}
\label{fig:Question Generation}
\vspace{-1em}
\end{figure*}
\vspace{-0.5em}
\subsection{Evaluation Dataset}
\label{sec:Evaluation Dataset}
\vspace{-0.5em}
\paragraph{Python Modules Selection}
We carefully choose modules that pertain to the fields of data science. It encompasses a wide array of topics such as data cleansing, visualization, image analysis, mathematical computations, and statistical methods. The modules we have selected are detailed in Tab.\ref{tab:python_modules}.


\vspace{-0.5em}
\paragraph{Topic Candidates Generation}
After selecting modules, we prompt (Appendix \ref{app: Construction Prompts}) GPT-4 to summarize 50 topics for each module, to encapsulate the vast majority of the module’s functionalities, thereby offering precise guidance for the subsequent generation of more targeted questions. 


\vspace{-0.5em}
\paragraph{Tasks Generation and Refinement}
We sample a module and topic, then prompt GPT-4 to generate questions and code based on the prompt in Fig.\ref{fig:Question Generation}. The prompt is designed to enable GPT-4 to generate a Jupyter notebook with sequential steps and various outputs, including numeric answers, structured output, and visualizations, mirroring real-world scenarios. Despite our request for concise descriptions, the generated content may lack conciseness and continuity. To address this, we undertake iterative refinement of these tasks. This process involves presenting both good and bad cases, along with additional modifications, to enhance the quality of questions and reduce ambiguity. Details of prompts used for refinement are in Appendix \ref{app: Construction Prompts}.

\vspace{-0.5em}
\paragraph{Human Refinement}
Despite the excellent capabilities of LLMs, we found that the generated tasks often focus on typical datasets, such as Titanic and Iris, which limits task diversity. Additionally, ensuring perfect accuracy in question descriptions and result correctness is challenging. 

To enhance diversity, we summarize template tasks based on the generated tasks and existing high-quality tutorials available in various Python libraries (Fig.\ref{fig:exp1}). With minor modifications, these template tasks can be adapted to a wide range of datasets. To further diversify the benchmark, we collect new datasets through two approaches: 1) We prompt GPT-4 to generate datasets tailored to different templates, leveraging GPT-4's ability to flexibly specify characteristics and data attributes. 2) We incorporate the latest datasets from the past year, ensuring authenticity and diversity, and significantly reducing the likelihood that the model has previously encountered this data.

To improve quality, we employ several experts who use a code interpreter for manual double-checking. This approach ensures that questions are written by real users and that various factors, such as runtime, are carefully controlled. For a comprehensive overview of the rules governing quality control, please refer to the Appendix \ref{app: qual-control}. The statistics of the dataset are also provided in the Appendix \ref{dataset_stats}.
\begin{figure*}[t]
    \centering
    \includegraphics[width=0.8\textwidth]{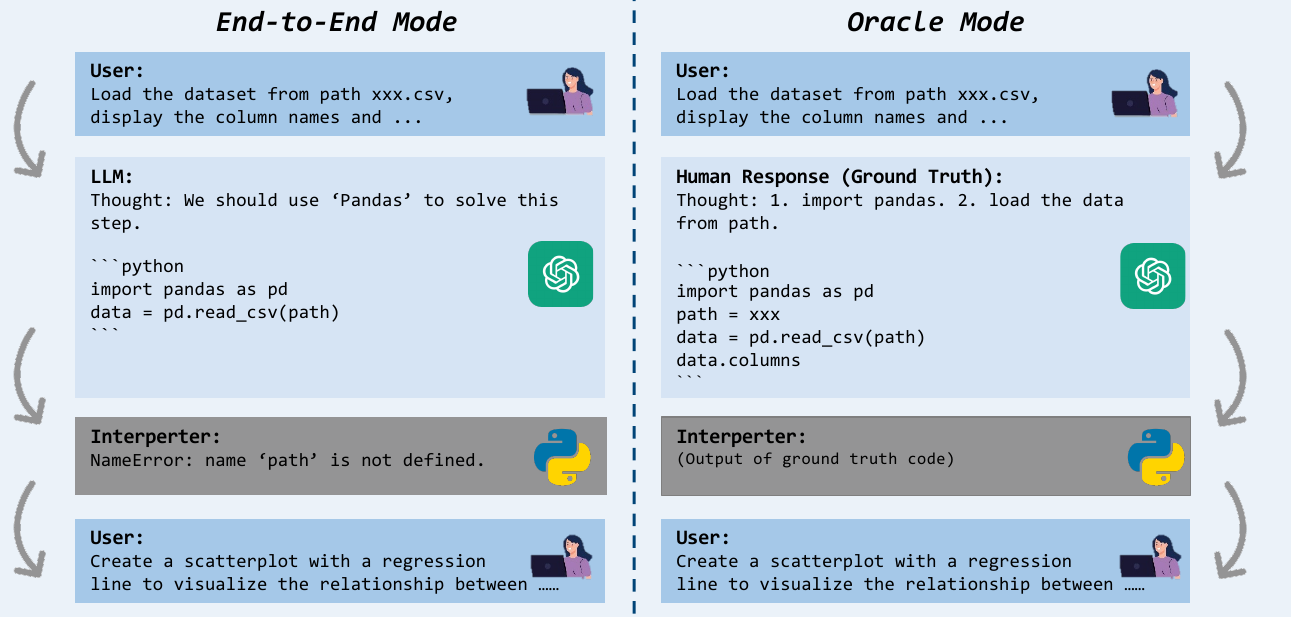}
    \vspace{-0.2em}
    \caption{Evaluation modes: In end-to-end mode, the LLM addresses the user's question (bottom) within the context of its response, while in oracle mode, it answers the user's question (bottom) within the context of ground truth.}
    \label{fig:full}
    \vspace{-1.8em}
\end{figure*}

\subsection{Evaluation Modes and Metrics}
\label{sec:Metrics}

\vspace{-0.5em}
\paragraph{Evaluation Modes} 
As shown in Fig.\ref{fig:full}, \shortname~ includes the end-to-end and oracle mode. It not only assesses the model's proficiency in autonomously resolving continuous problems but also assesses its capacity in conjunction with human interaction. In end-to-end mode, the model must solve the problem by itself. Each subsequent response is contingent upon the model's previous outcomes, necessitating self-correction based on code interpreter feedback. What's more, to reflect the real-world scenarios where human assistance is available, we introduce oracle mode to incorporate ground truth code and thought as context, helping the model to address subsequent tasks. Conceptually, oracle mode emulates a few-shot testing or in-context learning scenario, providing the model with comprehensive context to facilitate a more precise solution.

\vspace{-0.5em}
\paragraph{Evaluation Metrics}
In two evaluation modes, we introduce two types of evaluation metrics: process-oriented and output-oriented, to provide a comprehensive analysis of the model’s performance. Process-oriented metrics focus on the correct invocation of tools and the successful compilation and execution of code. These metrics include the \textit{Tool Call Rate}, which measures the proportion of instances where the model correctly follows the instructions to invoke a code interpreter, and the \textit{Executable Rate}, which indicates the percentage of code that is executed without any errors.

On the other hand, output-oriented metrics focus on the outcomes of the model. These metrics include \textit{Numeric Accuracy}, which assesses the accuracy of the numerical results; \textit{Text Score}, which measures the quality of the structural text output using the Rouge metric \cite{lin2004rouge}; and \textit{Visualization Score}, which evaluates the quality of visual output. Instead of using GPT-4V like Qwen-Agent, which is expensive and ineffective, we propose using structural similarities \cite{1284395} between predictions and ground truth images as the visualization score. These metrics provide a holistic evaluation of the LLM's capabilities.

\section{Experiments}
\vspace{-0.5em}
\subsection{Experiments Setup} \label{exp:detail}

\vspace{-0.5em}
To provide a thorough analysis, we evaluate 19 chat models, including popular open-sourced LLMs and the private GPT-4, using the \shortname~benchmark. During inference, we allow LLMs to attempt up to 3 times. The specific versions of Python modules utilized in the code interpreter are provided in Appendix \ref{module_version}. All experiments are conducted within the OpenCompass~\cite{2023opencompass} evaluation platform.

\newcommand\redcolor[1]{\cellcolor{gray!30!red!30}{#1}}
\newcommand\bluecolor[1]{\cellcolor{gray!30!blue!30}{#1}}
\begin{table*}[!t]
    \caption{\textbf{Main results of \shortname.} Tool, Exe, Num, Text, and Vis denote the tool call rate, executable rate, numeric accuracy, text score, and visualization score respectively. \textbf{bold} denotes the best score among the same model scale. \textit{Average is the mean of Num, Text, and Vis in two modes.}.}
    \label{tab:main_results}
    \setlength{\tabcolsep}{6pt}
    \renewcommand{\arraystretch}{1.1}
    \centering
    \resizebox{1.0\textwidth}{!}{
    \begin{tabular}{l|ccccc|ccccc|c}
    \toprule
    \multirow{2}{*}{\textbf{Model}} & \multicolumn{5}{c|}{\textbf{\textsc{End-to-end Mode}}} & \multicolumn{5}{c|}{\textbf{\textsc{Oracle Mode}}} & \multirow{2}{*}{\textcolor{orange}{Average}}\\
    \addlinespace[0.3pt]
    & \textcolor{olive}{\small{Tool}} & \textcolor{olive}{\small{Exe}} & \textcolor{orange}{\small{Num}} & \textcolor{orange}{\small{Text}} & \textcolor{orange}{\small{Vis}} & \textcolor{olive}{\small{Tool}} & \textcolor{olive}{\small{Exe}} & \textcolor{orange}{\small{Num}} & \textcolor{orange}{\small{Text}} & \textcolor{orange}{\small{Vis}}  \\ \midrule %
Llama-2-7B-Chat&   80.6   &   28.8   &   10.1   &   10.4   &   17.4   &   92.7   &   41.5   &   22.4   &   31.2   &   24.8   &   19.4   \\
Yi-6B-Chat&   90.6   &   50.2   &   22.5   &   21.7   &   31.1   &   99.7   &   61.5   &   29.5   &   40.8   &   40.2   &   31.0   \\
ChatGLM3-6B-32K&   77.9   &   48.8   &   18.9   &   33.0   &   29.3   &   98.6   &   55.7   &   28.2   &   52.0   &   34.2   &   32.6   \\
Vicuna-7B-v1.5-16k&   89.4   &   54.5   &   19.9   &   27.1   &   32.7   &   \underline{99.9}   &   62.9   &   29.3   &   43.8   &   43.1   &   32.7   \\
DeepSeek-7B-Chat&   62.9   &   45.3   &   16.0   &   17.4   &   30.8   &   93.8   &   67.1   &   33.0   &   56.7   &   48.5   &   33.7   \\
Qwen-7B-Chat&   \textbf{\underline{99.8}}   &   72.2   &   42.3   &   55.7   &   43.9   &   97.9   &   71.9   &   36.6   &   58.1   &   52.1   &   48.1   \\
Mistral-7b-Instruct&   97.5   &   67.2   &   41.9   &   53.6   &   41.7   &   99.2   &   75.7   &   48.0   &   55.4   &   49.1   &   48.3   \\
Qwen1.5-7B-Chat&   99.1   &   69.9   &   35.4   &   59.1   &   39.1   &   99.8   &   75.9   &   48.6   &   61.7   &   52.1   &   49.3   \\
InternLM2-7B-Chat&   99.7   &  77.7   &   52.6   &   64.6   &   51.8   &   99.3   &   70.4   &   48.1   &   44.3   &   42.9   &   50.7   \\
DeepSeek-Coder-6.7B&   75.9   &   66.0   &   40.7   &   48.5   &   47.7   &   96.8   &   83.0   &   59.2   &   77.2   &   \underline{62.5}   &   56.0   \\ 
Llama-3-8B-Instruct&   98.5   &   \underline{78.6}   &   \underline{55.6}   &   \underline{70.8}   &   \underline{54.6}   &   95.5   &   \underline{84.5}   &   \underline{60.4}   &   \underline{83.7}   &   61.4   &   \underline{64.4}   \\ \midrule
Llama-2-13B-Chat&   90.7   &   42.7   &   12.1   &   14.8   &   24.0   &   99.0   &   52.4   &   28.9   &   31.0   &   29.3   &   23.4   \\
Vicuna-13B-v1.5-16k&   91.0   &   62.2   &   25.9   &   45.4   &   35.4   &   99.9   &   72.1   &   39.7   &   45.2   &   45.9   &   39.6   \\
Qwen-14B-Chat&   93.1   &   70.9   &   48.2   &   60.7   &   48.7   &   99.2   &   82.4   &   55.5   &   74.6   &   57.3   &   57.5   \\
Qwen1.5-14B-Chat&   \underline{99.7}   &   \underline{84.8}   &   \underline{56.9}   &   63.3   &   51.4   &   99.9   &   \textbf{\underline{92.7}}   &   61.7   &   62.6   &   63.1   &   59.8   \\
Mixtral-8x7B-Instruct&   99.1   &   83.0   &   45.4   &   55.9   &   \underline{53.0}   &   \textbf{\underline{99.9}}   &   88.6   &   \underline{67.5}   &   73.5   &   \textbf{\underline{67.0}}   &   60.4   \\
InternLM2-20B-Chat&   96.6   &   82.2   &   47.5   &   \underline{67.2}   &   47.1   &   96.7   &   81.0   &   60.4   &   \textbf{\underline{85.0}}   &   56.7   &   \underline{60.6}   \\ \midrule
Yi-34B-Chat&   92.9   &   66.2   &   42.2   &   40.3   &   42.2   &   \underline{99.8}   &   77.2   &   68.2   &   72.2   &   52.7   &   53.0   \\
Llama-2-70B-Chat&   92.4   &   53.8   &   23.7   &   22.7   &   24.9   &   94.2   &   61.6   &   36.7   &   36.0   &   35.6   &   29.9   \\
DeepSeek-67B-Chat&   76.9   &   70.7   &   47.5   &   63.3   &   54.8   &   91.1   &   82.8   &   60.6   &   \underline{74.7}   &   63.8   &   60.8   \\
Qwen-72B-Chat&   \underline{98.3}   &   87.2   &   60.5   &   \textbf{\underline{73.2}}   &   55.9   &   99.7   &   \underline{90.1}   &   62.8   &   73.6   &   \underline{64.9}   &   65.1   \\
Llama-3-70B-Instruct&   98.0   &   \textbf{\underline{91.1}}   &   \textbf{\underline{64.9}}   &   55.7   &   \textbf{\underline{63.6}}   &   96.6   &   82.8   &   \textbf{\underline{72.9}}   &   74.2   &   62.0   &   \textbf{\underline{65.5}}   \\ \midrule
gpt-4o&   100.0   &   97.2   &   74.1   &   79.0   &   61.1   &   99.2   &   94.8   &   81.7   &   79.7   &   71.6   &   74.5   \\ 
GPT-4-1106-preview&   99.8   &   98.6   &   77.8   &   78.9   &   64.0   &   99.6   &   97.7   &   77.5   &   81.6   &   73.1   &   75.5   \\ \bottomrule
\end{tabular}
}
\vspace{-1em}
\end{table*}

\vspace{-0.5em}
\subsection{Main Results}
\label{sec:single_round}
\vspace{-0.5em}


We categorize the models into distinct groups based on their scales to enable more effective comparisons. As illustrated in Table \ref{tab:main_results}, LLama-3-8B-Instruct achieves the highest performance in the ~7B group, while InternLM2-20B-Chat demonstrates the top performance in the 13B-20B group. Within the 70B group, LLama-3-70B-Instruct secures the leading position with an overall score that is 10.0\% lower than GPT-4-1106-preview. The API model GPT-4-1106-preview and gpt-4o outperform all other models, especially in end-to-end mode, highlighting the significant potential for improvement in current open-source models.
What's more, larger models tend to exhibit superior performance across various metrics, in line with established trends \cite{brown2020language,kaplan2020scaling,wei2022emergent}. Moreover, models within the same series (such as Mistral, InternLM, Qwen, Llama3, etc.) consistently maintain relatively stable rankings within their respective parameter groups, underscoring the stability and efficacy of our approach.

When comparing the end-to-end mode and oracle mode, it becomes evident that the oracle mode surpasses the end-to-end mode across all metrics for most models. This observation suggests that LLMs can achieve better results with human interaction, hinting at a promising avenue for integrating LLMs to assist humans in data science. The experiment demos are shown in Appendix \ref{app: demo}.


\vspace{-0.5em}
\subsection{Error Mode Analysis}

\begin{figure*}[!t]
    \centering
    \includegraphics[width=0.9\textwidth]{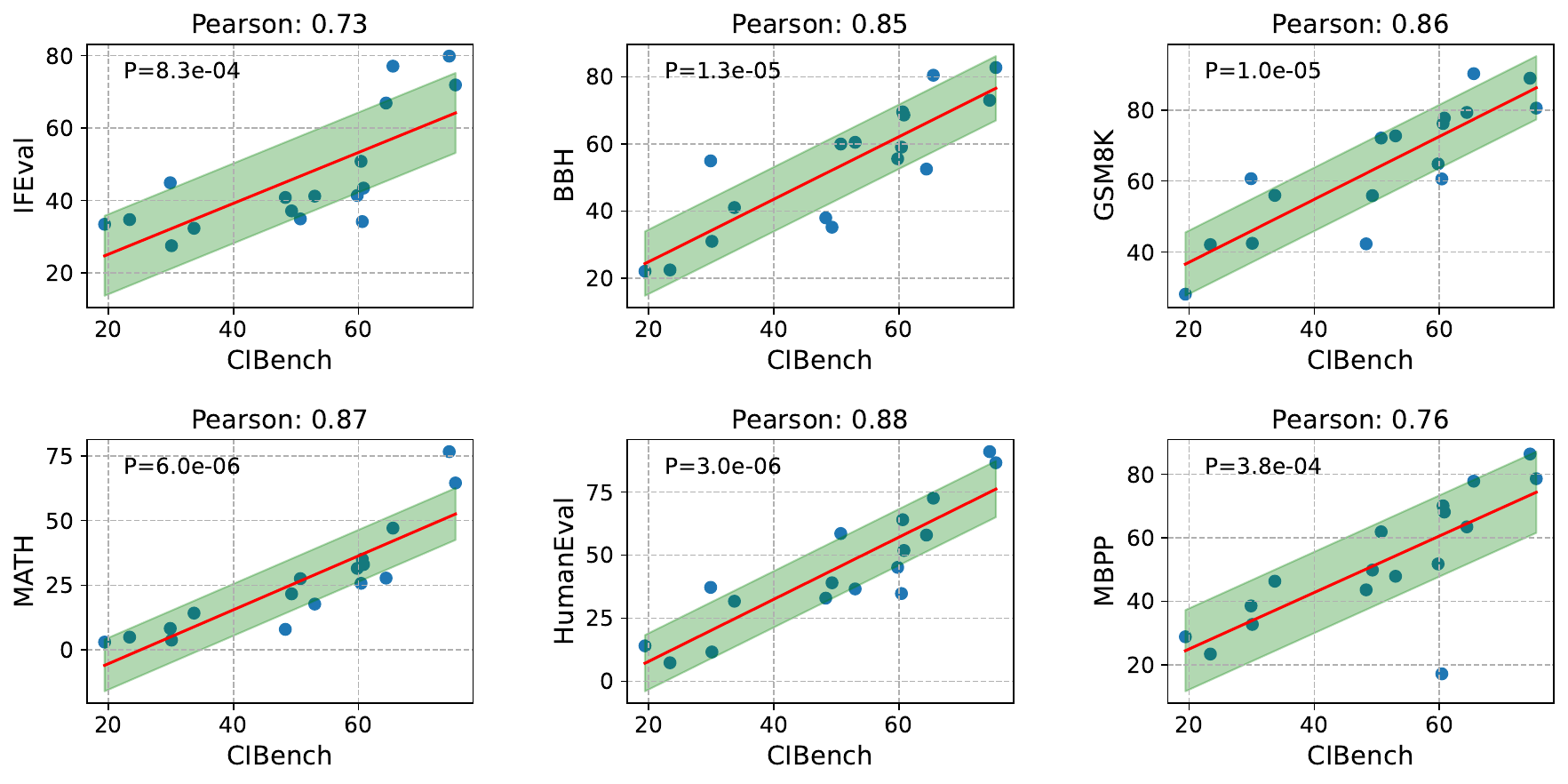}
    \vspace{-0.5em}
    \caption{Correlation of~\shortname~with other benchmarks. The small p-value (top-left) and high Pearson correlation coefficients (title) indicate a strong correlation between~\shortname~and IFEval, BBH, GSM8K, MATH, HumanEval, and MBPP. These benchmarks evaluate the instruction-following, reasoning, and coding abilities of LLMs, respectively.}
    \label{fig:reasoning}
    \vspace{-1.5em}
\end{figure*}

In the evaluation of \shortname, we identify four prevalent types of errors in the code generated by the model. These errors are categorized as follows:
1) \textit{Instruction Following Errors}: These encompass instances where the model deviates from or disregards provided instructions, reflecting a lack of adherence to specified guidelines; 2) \textit{Hallucination Errors}: This category pertains to the phenomenon of the model generating code that contains hallucinated elements, such as utilizing undefined parameters or referencing irrelevant variables; 3) \textit{Reasoning Errors}: These errors occur when the model encounters complex problems, often resulting in logical errors in the generated code. Such errors offer valuable insights into the model's ability to handle intricate tasks in code generation; 4) \textit{Code Errors}: Basic errors in code generation fall under this category. While these errors may sometimes appear trivial, they signify potential deficiencies in the model's code-generation process. These identified errors effectively underscore the current limitations of LLMs in terms of their coding capabilities, providing valuable insights for the ongoing development of \shortname. Since automatically classifying errors is challenging, we manually categorized the four types of errors in GPT-4-1106-preview. The proportions of these error types are 31.9\%, 4.3\%, 40.4\%, and 23.4\%, respectively. Detailed examples of these errors are presented in Appendix \ref{bad_case}.

Furthermore, to provide insight into enhancing performance on CIBench, we analyze its correlation with existing benchmarks such as IFEval, BBH, GSM8K, MATH, HumanEval, and MBPP. These benchmarks evaluate the instruction-following, reasoning, and coding abilities of LLMs. As shown in Fig.\ref{fig:reasoning}, CIBench performance exhibits a strong correlation with these benchmarks, with Pearson correlation coefficients exceeding 0.7, particularly for HumanEval and GSM8K. This result indicates that improvements in reasoning and coding can boost LLM performance on CIBench.

\subsection{More Analysis}
\begin{figure*}[!t]
    \centering
    \includegraphics[width=1.0\textwidth]{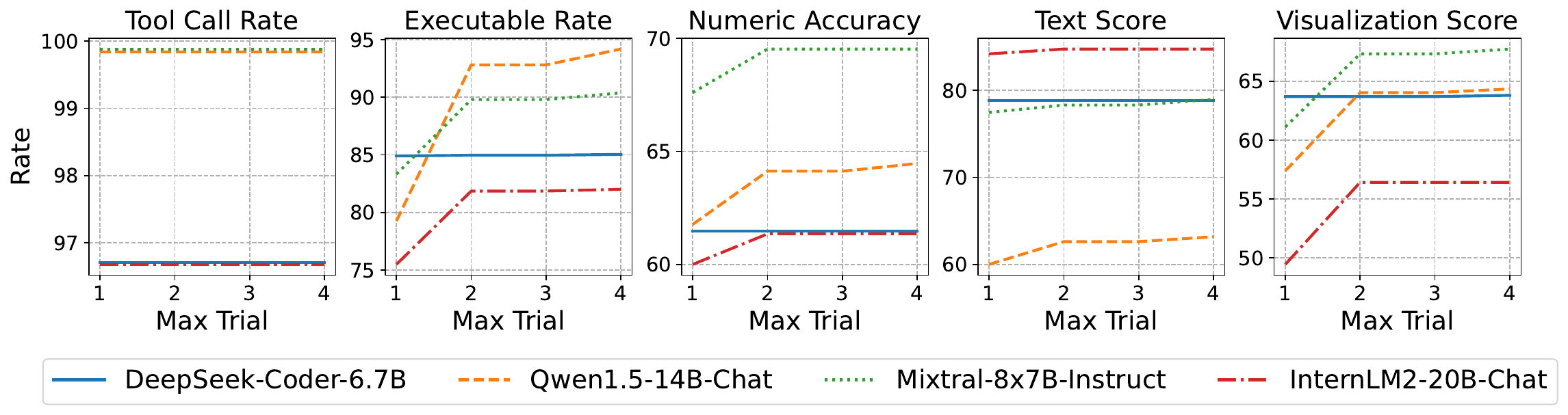}
    \caption{Debug ability analysis. Max trial denotes the trial times in ReAct protocol.}
    \label{fig:max_turn}
    \vspace{-1.5em}
\end{figure*}

\vspace{-0.5em}
\paragraph{Debug Ability Analysis}
In the ReAct protocol, we allow LLMs to try to solve tasks multiple times. During each trial, the model can use feedback from the code interpreter to rectify any bugs in the generated code. To assess the LLMs' ability to autonomously correct bugs, we vary the number of trials. As shown in Fig.\ref{fig:max_turn}, increasing the number of trials correlates with improvements across all metrics. Significantly, for most LLMs, there is a notable enhancement when the number of trials reaches two, particularly evident in metrics such as executable rate, numeric accuracy, and visualization score. 
This suggests that the LLM can autonomously rectify bugs to a certain extent. In our experiments, to balance evaluation time and performance, we set the number of trials to three.






\vspace{-0.5em}
\paragraph{Visualization Metric Analysis}
\label{sec:simple_vis}
To validate the effectiveness of our proposed visualization metric, we follow QwenAgent~\cite{bai2023qwen} and use GPT-4V to assess visualization scores on a subset of \shortname~tasks. The prompt is provided in Appendix \ref{app: sub-vis}.
As shown in Fig.\ref{fig:vis}, despite structural similarities being derived from low-level features, there is a strong correlation between them and GPT-4V scores, demonstrating remarkable consistency between the two metrics. Therefore, we can utilize structural similarities as a simplified visualization metric to subject GPT-4V for effective analysis. 

\vspace{-0.5em}
\paragraph{Cross Language Analysis}
\label{sec:cn_cibench}
To benchmark the LLMs' ability in Chinese, we created a Chinese version of \shortname~by translating the human-created tasks into Chinese. This allows us to evaluate the Code Interpreter performance in Chinese. As shown in Fig.\ref{fig:cn}, we observe that: 1) most models exhibit a slight decrease in Chinese \shortname~ compared to their English counterparts.; 2) the strong DeepSeek-67B-Chat and Qwen-72B-Chat drop a lot on Chinese ~\shortname, compared to the English version. Further research and development efforts are necessary to address these discrepancies and improve the performance of LLMs in multilingual scenarios.

\begin{figure}[!t]
\begin{minipage}{0.48\textwidth}
    \centering
    \includegraphics[width=0.8\textwidth]{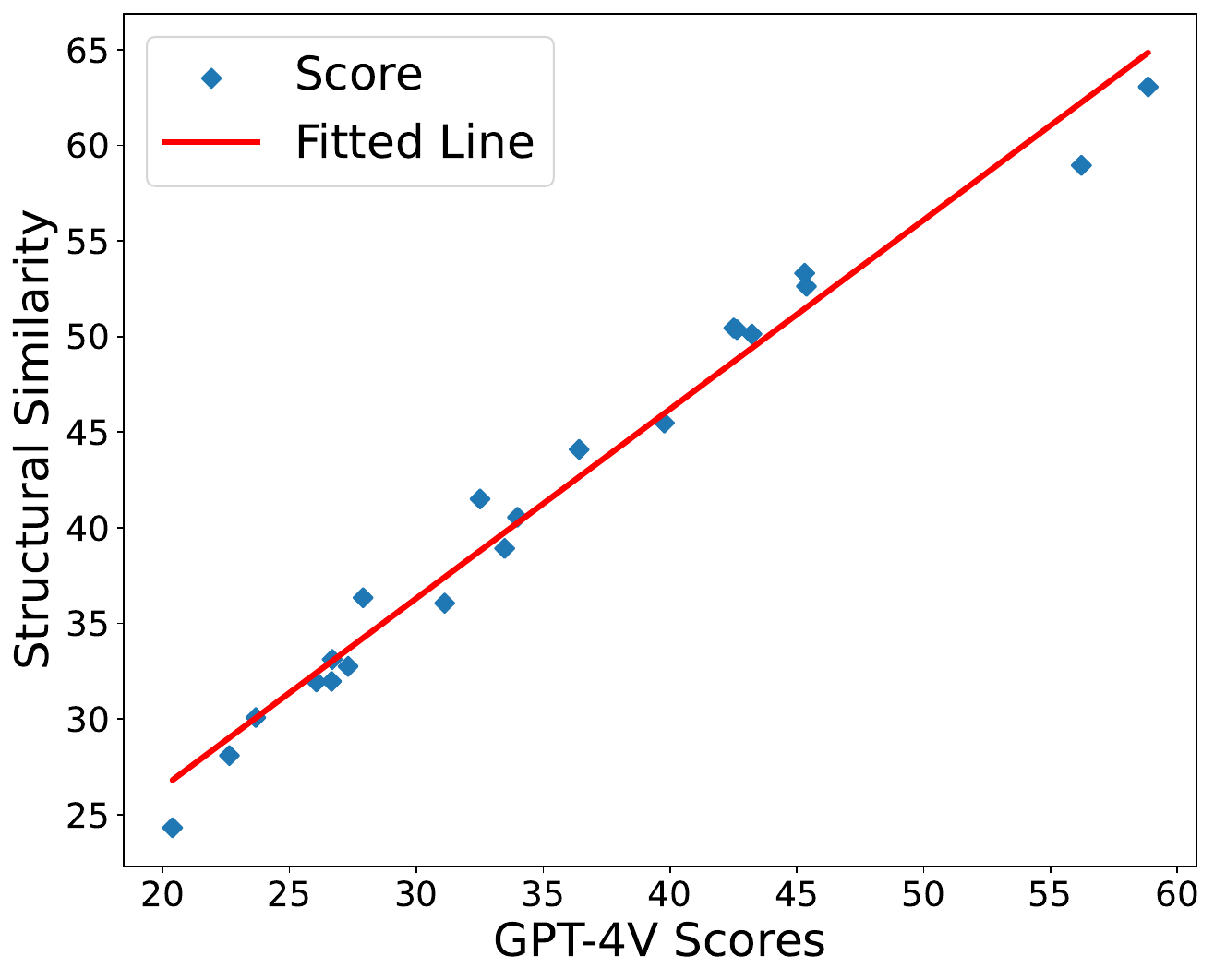}
    \caption{Structural Similarities v.s. GPT-4V.}
    \label{fig:vis}
    \vspace{-1em}
\end{minipage}
\begin{minipage}{0.49\textwidth}
    \centering
    \includegraphics[width=0.8\textwidth]{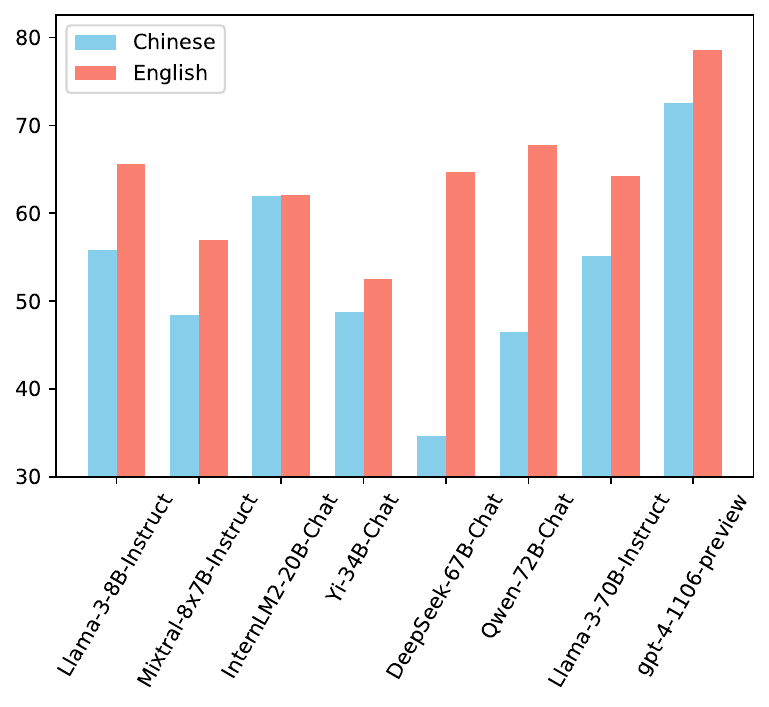}
    \caption{Chinese \shortname~ v.s. English \shortname~.}
    \label{fig:cn}
    \vspace{-1em}
\end{minipage}
\end{figure}

\begin{figure}[t]
\begin{minipage}{0.52\textwidth}
    \centering
    \small
    \begin{tabular}{c|ccc} \toprule
\textbf{Model} & \textbf{Easy} & \textbf{Medium} & \textbf{Hard} \\ \midrule
Yi-6B-chat & 37.5 & 33.1 & 23.1 \\
Llama-2-7B-chat & 25.7 & 19.2 & 15.7 \\
Qwen-7B-chat & 51.0 & 45.5 & 36.5 \\
Vicuna-7B-v1.5-16k & 39.0 & 35.9 & 23.8 \\
DeepSeek-7B-chat & 38.8 & 41.8 & 22.8 \\
Mistral-7B-instruct-v0.2 & 52.3 & 52.4 & 33.0 \\
Chatglm3-6B-32k & 43.2 & 33.3 & 19.0 \\
Mistral-8x7B-instruct-v0.1 & 62.1 & 61.2 & 47.6 \\
Internlm2-chat-7B & 55.5 & 52.0 & 33.9 \\
Qwen-14B-chat & 59.6 & 58.8 & 42.0 \\
Vicuna-13B-v1.5-16k & 46.1 & 41.6 & 27.7 \\
Llama-2-13B-chat & 34.8 & 23.5 & 18.7 \\
Internlm2-chat-20B & 51.5 & 59.2 & 45.1 \\
Yi-34b-chat & 55.2 & 58.1 & 38.0 \\
Llama-2-70B-chat & 38.1 & 33.9 & 18.6 \\
DeepSeek-67B-chat & 60.4 & 63.9 & 43.7 \\
Qwen-72B-chat & 60.7 & 66.2 & 50.3 \\ 
Llama-3-70B-chat & 62.3 & 67.1 & 42.0 \\\bottomrule
\end{tabular}
\caption{Performance on different difficulty levels. Easy means <2 steps, Medium means 2-4 steps, and Hard means >4 steps.}
    \label{tab:difficulty_level}

\end{minipage}\hfill 
\begin{minipage}{0.46\textwidth}
    \centering
    \includegraphics[width=1.0\textwidth]{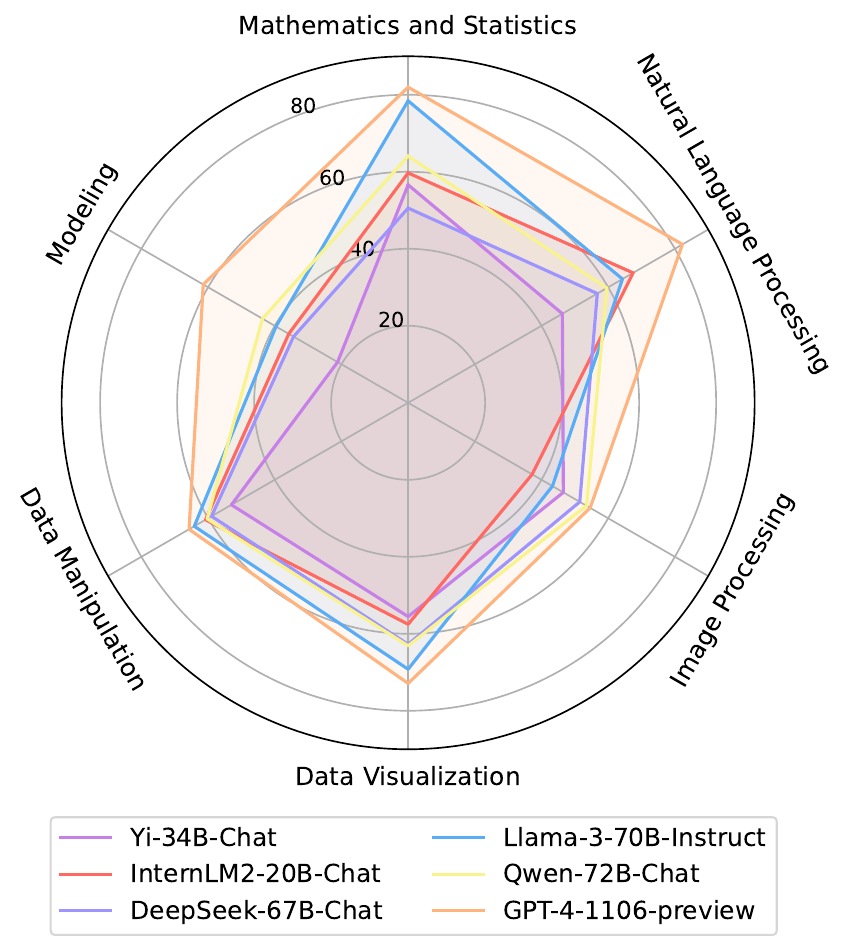}
    \caption{Performance on different categories.}
    \label{fig:radar}
\end{minipage}\hfill
\vspace{-1.5em}
\end{figure}

\vspace{-0.5em}
\paragraph{Difficulty Analysis} 
CIbench comprises interactive tasks, and we assume that varying interactive steps denote distinct levels of difficulty. To demonstrate model performance across these difficulty levels, we decompose each task into three categories: easy (up to 2 steps), medium (2-4 steps), and difficult (more than 4 steps), and then evaluate performance accordingly. As Tab.\ref{tab:difficulty_level} shows, for most models, the performance of most models decreases as the number of steps increases. 

\vspace{-0.5em}
\paragraph{Different Category Modules Analysis} 
We assess the capabilities of different LLMs with various category modules (refer to Tab. \ref{tab:python_modules}).
As shown in Fig. \ref{fig:radar}, LLMs demonstrate proficiency in tackling mathematical and statistical tasks utilizing SciPy modules. However, open-sourced LLMs encounter challenges when handling modeling tasks that demand advanced coding and reasoning skills. We hope that future open-sourced LLMs will excel in modeling tasks.
\vspace{-0.5em}
\paragraph{Limitation}\label{sec:limitation}
Our work has two main limitations: 1) \shortname~is currently limited to Python, despite it could be extended to include other programming languages using a similar methodology; 2) the evaluation metric of \shortname~has limitations in measuring certain data science tasks, such as "training a model with PyTorch" and tasks involving randomness.


\vspace{-0.8em}
\section{Conclusion}
\vspace{-0.8em}

We propose a novel benchmark, named~\shortname~, to comprehensively assess LLMs' ability to leverage code interpreters for complex data science tasks. It includes an evaluation dataset covering widely used Python modules in data science and two evaluation modes measuring LLMs' ability with and without human assistance. The evaluation dataset is constructed using an LLM-human cooperative approach, leveraging interactive IPython sessions to simulate realistic scenarios in data science. Thorough experimental analysis with 24 LLMs on \shortname~indicates that LLMs perform poorly in modeling category modules.
Based on our experimental analysis, we offer several insights for the future development of LLMs: \textit{enhancing the LLM's ability to correct errors based on feedback, improving its understanding of user intentions across multiple rounds of interactions, and, most importantly, strengthening its reasoning capabilities.}


\bibliography{neurips_2024}
\bibliographystyle{plain}
\section*{Checklist}

The checklist follows the references.  Please
read the checklist guidelines carefully for information on how to answer these
questions.  For each question, change the default \answerTODO{} to \answerYes{},
\answerNo{}, or \answerNA{}.  You are strongly encouraged to include a {\bf
justification to your answer}, either by referencing the appropriate section of
your paper or providing a brief inline description.  For example:
\begin{itemize}
  \item Did you include the license to the code and datasets? \answerYes{See }
  \item Did you include the license to the code and datasets? \answerNo{The code and the data are proprietary.}
  \item Did you include the license to the code and datasets? \answerNA{}
\end{itemize}
Please do not modify the questions and only use the provided macros for your
answers.  Note that the Checklist section does not count towards the page
limit.  In your paper, please delete this instructions block and only keep the
Checklist section heading above along with the questions/answers below.

\begin{enumerate}

\item For all authors...
\begin{enumerate}
  \item Do the main claims made in the abstract and introduction accurately reflect the paper's contributions and scope?
    \answerYes{See Introduction}
  \item Did you describe the limitations of your work?
    \answerYes{See Sec.\ref{sec:limitation}}
  \item Did you discuss any potential negative societal impacts of your work?
    \answerNA{Our paper proposes a dataset to measure the programming ability of large language models. There is currently no negative social impact.}
  \item Have you read the ethics review guidelines and ensured that your paper conforms to them?
    \answerYes{See Appendix}
\end{enumerate}

\item If you are including theoretical results...
\begin{enumerate}
  \item Did you state the full set of assumptions of all theoretical results?
    \answerNA{}
	\item Did you include complete proofs of all theoretical results?
    \answerNA{}
\end{enumerate}

\item If you ran experiments (e.g. for benchmarks)...
\begin{enumerate}
  \item Did you include the code, data, and instructions needed to reproduce the main experimental results (either in the supplemental material or as a URL)?
    \answerYes{\href{https://github.com/open-compass/CIBench}{https://github.com/open-compass/CIBench}}
  \item Did you specify all the training details (e.g., data splits, hyperparameters, how they were chosen)? \answerYes{\href{https://github.com/open-compass/CIBench}{https://github.com/open-compass/CIBench}}
	\item Did you report error bars (e.g., with respect to the random seed after running experiments multiple times)?
    \answerYes{The experiments are conducted at a temperature of 0.}
	\item Did you include the total amount of compute and the type of resources used (e.g., type of GPUs, internal cluster, or cloud provider)? \answerYes{See \ref{exp:detail}}
\end{enumerate}

\item If you are using existing assets (e.g., code, data, models) or curating/releasing new assets...
\begin{enumerate}
  \item If your work uses existing assets, did you cite the creators?
     \answerNA{}
  \item Did you mention the license of the assets?
     \answerNA{}
  \item Did you include any new assets either in the supplemental material or as a URL?
     \answerNA{}
  \item Did you discuss whether and how consent was obtained from people whose data you're using/curating?
     \answerNA{}
  \item Did you discuss whether the data you are using/curating contains personally identifiable information or offensive content?
     \answerNA{}
\end{enumerate}

\item If you used crowdsourcing or conducted research with human subjects...
\begin{enumerate}
  \item Did you include the full text of instructions given to participants and screenshots, if applicable?
    \answerYes{See Appendix}
  \item Did you describe any potential participant risks, with links to Institutional Review Board (IRB) approvals, if applicable?
   \answerNA{}
  \item Did you include the estimated hourly wage paid to participants and the total amount spent on participant compensation?
    \answerNA{}
\end{enumerate}

\end{enumerate}
\appendix
\definecolor{lightgray}{gray}{0.95}
\definecolor{codegreen}{rgb}{0,0.6,0}
\definecolor{codegray}{rgb}{0.5,0.5,0.5}
\definecolor{codepurple}{rgb}{0.58,0,0.82}
\lstdefinestyle{prompt}{
    basicstyle=\ttfamily\fontsize{7pt}{8pt}\selectfont,
    frame=none,
    breaklines=true,
    backgroundcolor=\color{lightgray},
    breakatwhitespace=true,
    breakindent=0pt,
    escapeinside={(*@}{@*)},
    numbers=none,
    numbersep=5pt,
    xleftmargin=5pt,
}
\lstdefinestyle{code}{
    language=Python,
    basicstyle=\ttfamily\fontsize{7pt}{8pt}\selectfont,
    keywordstyle=\color{magenta},
    commentstyle=\color{codegreen},
    stringstyle=\color{codepurple},
    showstringspaces=false，
    breaklines=true,
    backgroundcolor=\color{lightgray},
    breakatwhitespace=true,
    breakindent=0pt,
    numbers=none,
    numbersep=5pt,
    xleftmargin=5pt,
}
\tcbset{
  aibox/.style={
    top=10pt,
    colback=white,
    colframe=black,
    colbacktitle=black,
    enhanced,
    center,
    attach boxed title to top left={yshift=-0.1in,xshift=0.15in},
    boxed title style={boxrule=0pt,colframe=white,},
  }
}

\section{Dataset Details}
\label{app: Dataset Statistics}

\subsection{Module Version Settings}\label{module_version}
The version of Python modules used in code interpreters is listed in Tab.\ref{tab:module version}.

\begin{table}[!ht]
\small
\caption{The module version settings in \shortname.}
\label{tab:module version}
\centering
\begin{tabular}{l|l}
    \toprule
        \textbf{Module} & \textbf{Version} \\
    \midrule
    Pandas & 1.5.3\\
    Matplotlib & 3.7.2 \\
    Seaborn & 0.13.0 \\
    Scikit-learn & 1.2.1 \\
    PyTorch & 1.13.1 \\
    TensorFlow & 2.14.0 \\
    LightGBM & 4.1.0 \\
    NLTK & 3.8 \\
    PyTorch & 1.131 \\
    OpenCV-Python & 4.8.1.78 \\
    SciPy & 1.11.2 \\ \bottomrule
\end{tabular}
\end{table}

\subsection{Dataset Statistics}\label{dataset_stats}
The \shortname~comprises generation tasks, template tasks, and Chinese template tasks, which produce three types of output: numerical, text, and visualization. The statistics of \shortname~are shown in Tab.\ref{tab:data stats}.

\begin{table}[!ht]
\small
\caption{Dataset statistics of \shortname. "generation", "template", and "template\_cn" represent generation tasks, template tasks, and Chinese template tasks, respectively. \textbf{Other} refers to the questions that only require successful execution without any output comparison.}
\label{tab:data stats}
\centering
\begin{tabular}{c|cccc|c}
    \toprule
        \textbf{Subset} & \textbf{Num} & \textbf{Text} & \textbf{Vis} & \textbf{Other} & \textbf{Total} \\
    \midrule
        generation & 210 & 76 & 466 & 208 & 960 \\ 
        template & 147 & 20 & 161 & 142 & 470 \\ 
        template\_cn & 147 & 20 & 161 & 142 & 470 \\ 
    \midrule
        total & 504 & 116 & 788 & 492 & 1900 \\ 
    \bottomrule
\end{tabular}
\end{table}

\section{Construction Prompts and Rules}
\label{app: Construction Prompts}

\paragraph{Topic Generation}
\label{app: topic-gen}
The prompt used for topic generation is shown in Fig. \ref{fig: Topic Generation}.
\paragraph{Question Refinement}
\label{app: exp-refine}
The prompts used for question refinement are shown in Fig. \ref{fig: Question Refinement}, \ref{fig: Another Question Refinement}.
\paragraph{Quality Control Rules}
\label{app: qual-control}
We include manual quality checking in the final steps due to limitations in the LLMs' ability to control runtime or file size. The rules used for quality control are shown in Fig.\ref{fig: Quality Control Rules}.
\section{Experiment Example Demo}
\label{app: demo}
\begin{figure*}[!t]
    \centering
    \includegraphics[width=1.0\columnwidth]{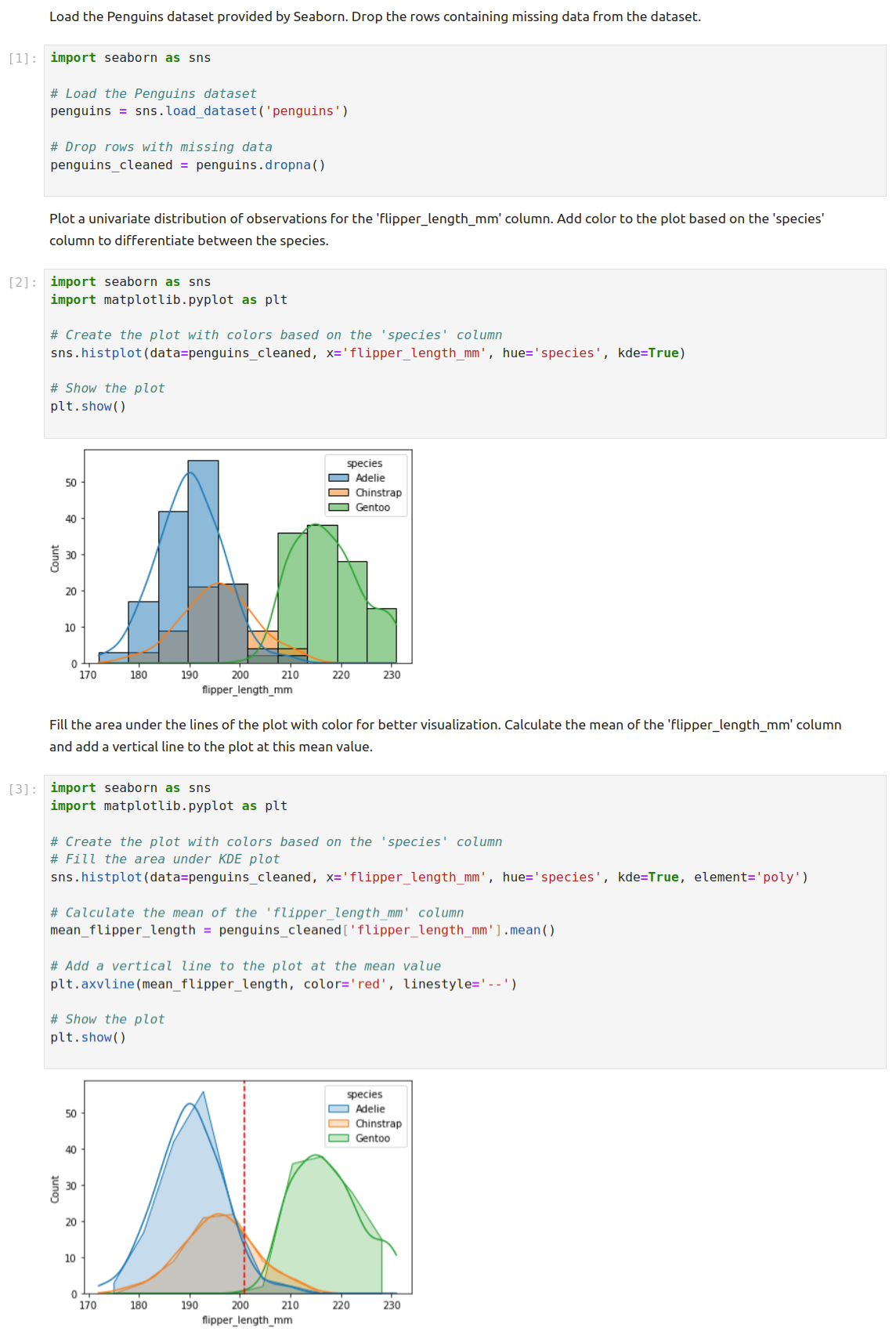}
    \caption{Example of successful template task. The penguins dataset and the specific attribute "flipper\_length\_mm" can be substituted with alternative datasets and columns. Such a template task enriches the diversity of~\shortname.}
    \label{fig:exp1}
    \vspace{-2.5mm}
\end{figure*}

\begin{figure*}[!t]
    \centering
    \includegraphics[width=1.0\columnwidth]{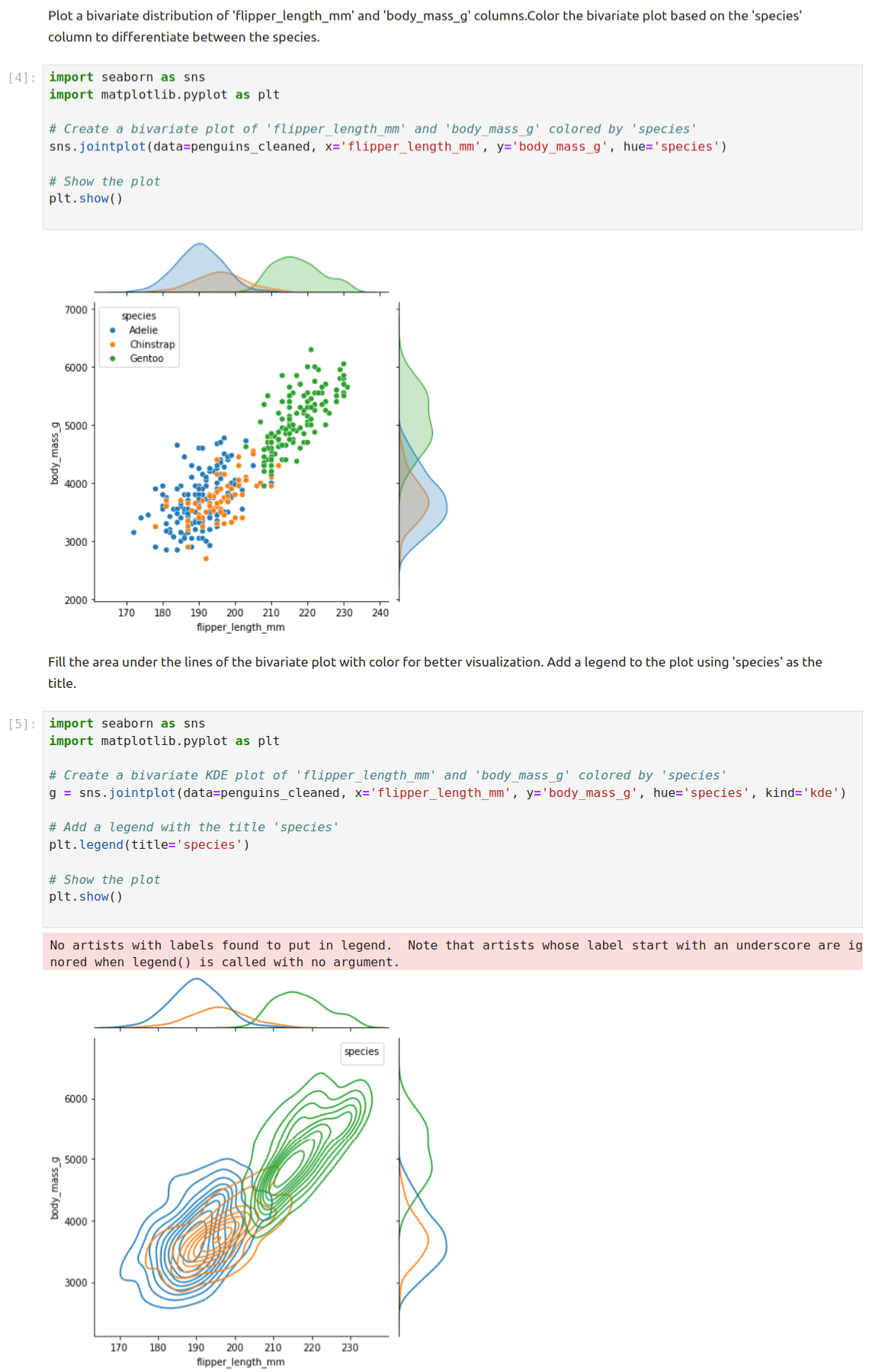}
    \caption{Example of successful template task.  Continue of Fig. \ref{fig:exp1}.}
    \label{fig:exp2}
    \vspace{-2.5mm}
\end{figure*}

An example of the experiment is shown in Fig. \ref{fig:exp1} and Fig. \ref{fig:exp2}.

\section{Subjective Visualization Evaluation}
\label{app: sub-vis}
The prompt used for subjective visualization scoring is shown in Fig.\ref{fig: sub-vis-prompt}.

\section{Dataset Error Analysis}\label{bad_case}

In the evaluation of \shortname, we identify four prevalent types of errors in the code generated by the model. These errors are categorized as follows:
1) \textbf{Instruction Following Errors} (Fig.\ref{fig: Following error}): These encompass instances where the model deviates from or disregards provided instructions, reflecting a lack of adherence to specified guidelines; 2) \textbf{Hallucination Errors} (Fig.\ref{fig: Hallucination error}): This category pertains to the phenomenon of the model generating code that contains hallucinated elements, such as utilizing undefined parameters or referencing irrelevant variables; 3) \textbf{Reasoning Errors} (Fig.\ref{fig: Reasoning error}): These errors occur when the model encounters complex problems, often resulting in logical errors in the generated code. Such errors offer valuable insights into the model's ability to handle intricate tasks in code generation; 4) \textbf{Code Errors} (Fig.\ref{fig: Code error}): Basic errors in code generation fall under this category. While these errors may sometimes appear trivial, they signify potential deficiencies in the model's code-generation process. These identified errors effectively underscore the current limitations of LLMs in terms of their Code Interpreter capabilities, providing valuable insights for the ongoing development of \shortname. 


\section{Human Annotator}

In our paper, the authors also serve as annotators for CIBench. These annotators are tasked with creating template tasks for various data science challenges, ensuring coverage of a wide range of concepts in the field. Importantly, our data collection process for these tasks is unbiased and does not involve any private information.


\section{Ethical Consideration}
We use GPT-4 and online resources to construct our benchmark. The benchmarks are carefully processed by experts to exclude any private information. Additionally, we utilize ChatGPT for text refinement and to correct any typographical errors during the writing process.

\begin{figure*}[!ht] 

\begin{AIbox}{Topic Generation}
{\color{blue}\bf Prompt:} \\
{
Please create 50 different topics base on Python module \{\}.
These topics can cover the majority of functionality of this module.
}
\end{AIbox} 
\caption{An example prompt of topic generation.}
\label{fig: Topic Generation}

\end{figure*}

%
%

\begin{figure*}[!ht] 
\begin{AIbox}{Question Refinement}
{\color{blue}\bf Prompt:} \\
{
Based on the experiment document and each step, please refine these steps into specific instructions for code writing to solve a real world question.\\
Provide parameter settings if needed, ensuring that the execution results are identical for any reasonable person attempting to write the code.\\
Please refrain from directly mentioning the name of the API, just target on purpose and give parameter settings.\\
Please also summarize the required external dataset if applicable.\\

Bad case step:\\

1. "Remove the 'Cabin' column from the dataframe due to its numerous missing values. Make sure to specify that you are removing a column and that the change should be permanent."\\
Too wordy, "Remove the 'Cabin' column and make the change permanent." should be fine.\\

2. "Identify and count how many values are missing in each column of the dataframe. Use a chain of two functions: the first checks if each value is null, and the second adds up these null instances for each column."\\
Too specific, "Count how many values are missing in each column of the dataframe. Print the count." should be fine.\\

Good case step:\\

1. "Convert the 'Sex' column to numerical values. Map 'male' to 0 and 'female' to 1."
Concise, clear instruction.\\

Your output format should be as follows starting with import necessary libraries:\\
$\left[Optional\right]$ Dataset Link:\\
$\left[Optional\right]$ Dataset Description:\\

Step 1. xx\\
\# code blcok\\
Step 2. xx\\
\# code blcok\\
...
}
\end{AIbox} 
\caption{An example prompt of question refinement.}
\label{fig: Question Refinement}
\end{figure*}

\begin{figure*}[!ht] 

\begin{AIbox}{Another Question Refinement}
{\color{blue}\bf Prompt:} \\
{
Given the above experiment document. Do the following modification:

1. Remove all the installation steps.\\
2. Remove all the file saving steps.\\
3. Elaborate the steps to be more specific with number and inputs that the execution results are identical for any reasonable person attempting to solve this step.\\
4. Reorder the steps.\\

Response with the new experiment document.
}
\end{AIbox} 
\caption{Another example prompt of question refinement.}
\label{fig: Another Question Refinement}

\end{figure*}

\begin{figure*}[!ht] 
\begin{AIbox}{Quality Control Rules}
{\color{blue}\bf Check Rules:} 
{\begin{itemize}
    \item Assure the questions is written from the perspective of a real user.
    \item Assure file path informed in the head for all the experiment required external files.
    \item Control the runtime, and each step should ideally produce results within 1 minute.
    \item Control the file size, the file used for single experiment should ideally not exceed 50M.
    \item Assure the output is valid and unambiguous as ground truth.
\end{itemize}
}
\end{AIbox} 
\caption{Rules used for manual dataset quality control.}
\label{fig: Quality Control Rules}
\end{figure*}

\begin{figure*}[!ht] 
\begin{AIbox}{Subjective Visualization Scoring Prompt}
{\color{blue}\bf Prompt:} \\
{You are an assistant skilled in assessing visualization capabilities.\\
In the capacity of a fair judge, you will evaluate the quality of images drawn by an AI model generating code for visualization-related problems. We will provide you with a code visualization problem and an image drawn by the code created by the AI model you need to assess. When you start your assessment, you must adhere to the following process:\\
1. Rate the visualization capability with a score between 1 and 10 for the image, returning only the number without any additional descriptions.\\
2. Your scoring needs to be as rigorous as possible, and it should follow the scoring rules below: Overall, the higher the quality of the model's response, the higher the score.\\

A score of 1 to 2 is given when the image cannot reflect the given instruction content at all.\\
A score of 3 to 4 is given when the image can partly reflect the given instruction content, but there is a significant lack of specific detail expression.\\
If the image basically meets the given instructions, but the aesthetic quality of the image is average without any outstanding features, this kind of rating can get a score of 5 to 6.\\
When the image matches the given instructions well, and shows some aesthetic appeal, such as some originality in color matching and shape design, this kind of rating can get a score of 7 to 8.\\
Only when the image completely matches the given instructions, covers all the details in the instructions, and performs excellently in terms of aesthetics, can this kind of rating get a score of 9 to 10.\\

[Visualization Problem]:\{question\}
}
\end{AIbox} 
\caption{An example prompt of subjective visualization scoring prompt.}
\label{fig: sub-vis-prompt}
\end{figure*}

\begin{figure*}[!ht] 

\begin{AIbox}{Following error}
{\color{orange}\bf User:} \\
{
Detect corners using the Shi-Tomas corner detector with max-corners=50, min-distance=0.5 and blocksize=10, mark the corners with circles on the image. The radius and thickness of the circle are 5 and 1. Show the marked image.Please use opencv and numpy modules.
}\\
{\color{teal}\bf Assistant:}
\begin{lstlisting}[style=code]
# Detect corners using Shi-Tomasi corner detector
corners = cv2.goodFeaturesToTrack(equalized_image, 
                                  maxCorners=max_corners, 
                                  qualityLevel=quality_level, 
                                  minDistance=min_distance, 
                                  blockSize=block_size)
# Mark the corners with circles on the image
marked_image = equalized_image.copy()
for corner in corners:
    x, y = corner.ravel()
    cv2.circle(marked_image, (int(x), int(y)), 5, (255, 0, 0), -1)  # Draw a blue filled circle at each corner

# Show the marked image
plt.imshow(marked_image, cmap='gray')
plt.axis('off')  # Turn off axis numbers and ticks
plt.show()

\end{lstlisting}
{\color{blue}\bf System:}
\begin{lstlisting}[style=prompt]
NameError                                 Traceback (most recent call last)
Cell In[6], line 2
      1 # Detect corners using Shi-Tomasi corner detector
----> 2 corners = cv2.goodFeaturesToTrack(equalized_image, maxCorners=max_corners, qualityLevel=quality_level, minDistance=min_distance, blockSize=block_size)
      4 # Mark the corners with circles on the image
      5 marked_image = equalized_image.copy()

NameError: name 'max_corners' is not defined
\end{lstlisting}
\hrulefill

{\color{orange}\bf User:}
{
Define the CrossEntropyLoss function and the Adam optimizer. Train the model for 3 epochs. Display the loss for the last epoch(Keep to two decimal places).
}\\
{\color{teal}\bf Assistant:}
\begin{lstlisting}[style=code]
import torch.optim as optim

criterion = nn.CrossEntropyLoss()
optimizer = optim.Adam(model.parameters(), lr=0.001)

epochs = 3
for epoch in range(epochs):
    model.train()
    running_loss = 0.0
    for inputs, labels in train_loader:
        inputs, labels = inputs.to(device), labels.to(device)
        
        optimizer.zero_grad()
        
        outputs = model(inputs)
        loss = criterion(outputs.squeeze(), labels.float())
        loss.backward()
        optimizer.step()
        
        running_loss += loss.item()
    
    final_loss = round(running_loss/len(train_loader), 2)
    print(final_loss)
\end{lstlisting}

{\color{blue}\bf System:}
\begin{lstlisting}[style=prompt]
933.55
933.45
933.09
\end{lstlisting}

\end{AIbox} 
\caption{Examples of Following error}
\label{fig: Following error}

\end{figure*}

\begin{figure*}[!ht] 

\begin{AIbox}{Hallucination error}
{\color{orange}\bf User:} \\
{
Build a vocabulary of all the tokens and assign an index to the vocabulary. Display the index of the word "happy". Please use nltk modules.
}\\
{\color{teal}\bf Assistant:}
\begin{lstlisting}[style=code]
# Check the vocabulary
vocab
# Display the index of the word "happy"
word_to_index.get('happy', 'Word not found in vocabulary')
\end{lstlisting}

{\color{blue}\bf System:}
\begin{lstlisting}[style=prompt]
NameError                                 Traceback (most recent call last)
Cell In[2], line 2
      1 # Check the vocabulary
----> 2 vocab
      3 # Display the index of the word "happy"
      4 word_to_index.get('happy', 'Word not found in vocabulary')

NameError: name 'vocab' is not defined
\end{lstlisting}

\hrulefill

{\color{orange}\bf User:} \\
{
Define a LightGBM model  with max\_depth=4, n\_estimators=120, learning\_rate=0.01 and num\_leaves=31. Train the model with Evaluation Metric='logloss'. Please use LightGBM modules.
}\\
{\color{teal}\bf Assistant:}
\begin{lstlisting}[style=code]
import lightgbm as lgb

# Define the LightGBM model
model = lgb.LGBMClassifier(max_depth=4, n_estimators=120, learning_rate=0.01, num_leaves=31)

# Train the model with evaluation metric as 'logloss'
model.fit(X_train, y_train, eval_set=(X_test, y_test), eval_metric='logloss', early_stopping_rounds=20)
\end{lstlisting}

{\color{blue}\bf System:}
\begin{lstlisting}[style=prompt]
TypeError                                 Traceback (most recent call last)
Cell In[3], line 7
      4 model = lgb.LGBMClassifier(max_depth=4, n_estimators=120, learning_rate=0.01, num_leaves=31)
      6 # Train the model with evaluation metric as 'logloss'
----> 7 model.fit(X_train, y_train, eval_set=(X_test, y_test), eval_metric='logloss', early_stopping_rounds=20)

TypeError: LGBMClassifier.fit() got an unexpected keyword argument 'early_stopping_rounds'
\end{lstlisting}

\end{AIbox} 
\caption{Examples of Hallucination error}
\label{fig: Hallucination error}

\end{figure*}

\begin{figure*}[!ht] 
\vspace{-10mm}
\begin{AIbox}{Reasoning error}
{\color{orange}\bf User:} \\
{
Split the dataset into training and testing sets using 0.2 as the test size, then define the train\_loader and test\_loader. Set batch size as 64. Please use pytorch and sklearn modules.
}\\
{\color{teal}\bf Assistant:}
\begin{lstlisting}[style=code]
from sklearn.model_selection import train_test_split
import torch
from torch.utils.data import DataLoader, TensorDataset

# Split the dataset into training and testing sets
X = df['text'].values
y = df['sentiment'].values
X_train, X_test, y_train, y_test = train_test_split(X, y, test_size=0.2, random_state=42)

# Define the train_loader and test_loader
def create_data_loader(X, y, batch_size):
    X_tensor = torch.tensor(X)
    y_tensor = torch.tensor(y, dtype=torch.long)
    dataset = TensorDataset(X_tensor, y_tensor)
    data_loader = DataLoader(dataset, batch_size=batch_size)
    return data_loader

batch_size = 64
train_loader = create_data_loader(X_train, y_train, batch_size)
test_loader = create_data_loader(X_test, y_test, batch_size)
\end{lstlisting}

{\color{blue}\bf System:}
\begin{lstlisting}[style=prompt]
TypeError                                 Traceback (most recent call last)
Cell In[4], line 19
     16     return data_loader
     18 batch_size = 64
---> 19 train_loader = create_data_loader(X_train, y_train, batch_size)
     20 test_loader = create_data_loader(X_test, y_test, batch_size)

Cell In[4], line 12
     11 def create_data_loader(X, y, batch_size):
---> 12     X_tensor = torch.tensor(X)
     13     y_tensor = torch.tensor(y, dtype=torch.long)
     14     dataset = TensorDataset(X_tensor, y_tensor)

TypeError: can't convert np.ndarray of type numpy.object_. The only supported types are: float64, float32, float16, complex64, complex128, int64, int32, int16, int8, uint8, and bool.
\end{lstlisting}

\hrulefill

{\color{orange}\bf User:} \\
{
Group and aggregate data by weight and calculate the average of each numerical group. Find out how many attributes have a negative correlation with weight.Please use pandas modules.
}\\
{\color{teal}\bf Assistant:}
\begin{lstlisting}[style=code]
# Group and aggregate data by weight and calculate the average of each numerical group
grouped_data = df.groupby('Weight').mean()

# Find attributes with negative correlation with weight
negative_correlation_count = (grouped_data.corr()['Weight'] < 0).sum()
negative_correlation_count
\end{lstlisting}
{\color{blue}\bf System:}
\begin{lstlisting}[style=prompt]
KeyError                                  Traceback (most recent call last)
File anaconda3/envs/prod-torch1.13/lib/python3.10/site-packages/pandas/core/indexes/base.py:3802, in Index.get_loc(self, key, method, tolerance)
   3801 try:
-> 3802     return self._engine.get_loc(casted_key)
   3803 except KeyError as err:

File anaconda3/envs/prod-torch1.13/lib/python3.10/site-packages/pandas/_libs/index.pyx:138, in pandas._libs.index.IndexEngine.get_loc()

File anaconda3/envs/prod-torch1.13/lib/python3.10/site-packages/pandas/_libs/index.pyx:165, in pandas._libs.index.IndexEngine.get_loc()

File pandas/_libs/hashtable_class_helper.pxi:5745, in pandas._libs.hashtable.PyObjectHashTable.get_item()

File pandas/_libs/hashtable_class_helper.pxi:5753, in pandas._libs.hashtable.PyObjectHashTable.get_item()

KeyError: 'Weight'
\end{lstlisting}

\end{AIbox} 
\caption{Examples of Reasoning error}
\label{fig: Reasoning error}

\end{figure*}

\begin{figure*}[!ht] 

\begin{AIbox}{Code error}
{\color{orange}\bf User:} \\
{
Generate a pair plot for all the numerical columns, using color = "pastel".Please use seaborn modules.
}\\
{\color{teal}\bf Assistant:}
\begin{lstlisting}[style=code]
sns.pairplot(df, kind='scatter', plot_kws={'color':'pastel'})
plt.show()
\end{lstlisting}
{\color{blue}\bf System:}
\begin{lstlisting}[style=prompt]
ValueError                                Traceback (most recent call last)
File anaconda3/envs/prod-torch1.13/lib/python3.10/site-packages/matplotlib/axes/_axes.py:4372, in Axes._parse_scatter_color_args(c, edgecolors, kwargs, xsize, get_next_color_func)
   4371 try:
-> 4372     mcolors.to_rgba_array(kwcolor)
   4373 except ValueError as err:

File anaconda3/envs/prod-torch1.13/lib/python3.10/site-packages/matplotlib/colors.py:471, in to_rgba_array(c, alpha)
    470 if isinstance(c, str):
--> 471     raise ValueError(f"{c!r} is not a valid color value.")
    473 if len(c) == 0:

ValueError: 'pastel' is not a valid color value.
\end{lstlisting}

\hrulefill

{\color{orange}\bf User:} \\
{
Blur the initial image by applying a Gaussian filter with a sigma value of 3. Sigma is the standard deviation for the Gaussian kernel, which controls the amount of blur. Then display the smoothed image.
}\\
{\color{teal}\bf Assistant:}
\begin{lstlisting}[style=code]
smooth = ndimage.gaussian_filter(image, sigma=3)
plt.imshow(smooth)
plt.show()
\end{lstlisting}

{\color{blue}\bf System:}
\begin{lstlisting}[style=prompt]
NameError                                 Traceback (most recent call last)
Cell In[48], line 2
      1 smooth = ndimage.gaussian_filter(image, sigma=3)
----> 2 plt.imshow(smooth)
      3 plt.show()

NameError: name 'plt' is not defined
\end{lstlisting}

\end{AIbox} 
\caption{Examples of Code error}
\label{fig: Code error}

\end{figure*}


%

\end{document}